\documentclass[sigconf]{aamas} 

\usepackage{balance} 
\usepackage{amsmath}
\usepackage{amsfonts}
\usepackage{subcaption}
\usepackage[ruled, vlined, linesnumbered]{algorithm2e}

\setcopyright{ifaamas}
\acmConference[AAMAS '21]{Proc.\@ of the 20th International Conference on Autonomous Agents and Multiagent Systems (AAMAS 2021)}{May 3--7, 2021}{Online}{U.~Endriss, A.~Now\'{e}, F.~Dignum, A.~Lomuscio (eds.)}
\copyrightyear{2021}
\acmYear{2021}
\acmDOI{}
\acmPrice{}
\acmISBN{}

\acmSubmissionID{145}

\SetKwInput{KwData}{Input}
\SetKwInput{KwResult}{Output}

\title[AAMAS-2021 Formatting Instructions]{Identification of Unexpected Decisions in Partially Observable Monte-Carlo Planning: A Rule-Based Approach}

\author {
    Giulio Mazzi
}
\affiliation{
    \department{Dipartimento di Informatica}
    \institution{Università degli Studi di Verona}
    \city{Verona}
    \state{Italy}
}
\email{giulio.mazzi@univr.it}

\author {
    Alberto Castellini
}
\affiliation{
    \department{Dipartimento di Informatica}
    \institution{Università degli Studi di Verona}
    \city{Verona}
    \state{Italy}
}
\email{alberto.castellini@univr.it}

\author {
    Alessandro Farinelli
}
\affiliation{
    \department{Dipartimento di Informatica}
    \institution{Università degli Studi di Verona}
    \city{Verona}
    \state{Italy}
}
\email{alessandro.farinelli@univr.it}

\begin{abstract}
Partially Observable Monte-Carlo Planning (POMCP) is a powerful online algorithm able to generate approximate policies for large Partially Observable Markov Decision Processes. The online nature of this method supports scalability by avoiding complete policy representation. The lack of an explicit representation however hinders interpretability. In this work, we propose a methodology based on Satisfiability Modulo Theory (SMT) for analyzing POMCP policies by inspecting their traces, namely sequences of belief-action-observation triplets generated by the algorithm. The proposed method explores local properties of policy behavior to identify unexpected decisions. We propose an iterative process of trace analysis consisting of three main steps, i) the definition of a \emph{question} by means of a parametric logical formula describing (probabilistic) relationships between beliefs and actions, ii) the generation of an \emph{answer} by computing the parameters of the logical formula that maximize the number of satisfied clauses (solving a MAX-SMT problem), iii) the analysis of the generated logical formula and the related decision boundaries for identifying unexpected decisions made by POMCP with respect to the original question. We evaluate our approach on Tiger, a standard benchmark for POMDPs, and a real-world problem related to mobile robot navigation. Results show that the approach can exploit human knowledge on the domain, outperforming state-of-the-art anomaly detection methods in identifying unexpected decisions. An improvement of the Area Under Curve up to $47\%$ has been achieved in our tests.

\end{abstract}


\DeclareRobustCommand{\freevar}[1]{\overline{\textbf{#1}}}

\usepackage{xcolor}

\definecolor{_orange}{RGB}{217,95,2}
\definecolor{_green}{RGB}{27,158,119}
\definecolor{_purple}{RGB}{117,112,179}
\definecolor{_blue}{RGB}{0, 114, 178}
\definecolor{_lightblue}{RGB}{86,180,233}

\begin{document}


\pagestyle{fancy}
\fancyhead{}


\maketitle 

\section{Introduction}
Planning in a partially observable environment is an important problem in artificial intelligence and robotics.
A popular framework to model such problem is \emph{Partially Observable Markov Decision Processes (POMDPs)}~\cite{Cassandra97} which encode dynamic systems where the state is not directly observable but must be inferred from observations.
Computing optimal policies, namely functions that map beliefs (i.e., probability distributions over states) to actions, in this context is PSPACE-complete~\cite{Papadimitriou1987}.
However, recent approximate and online methods allow handling many real-world problems.
A pioneering algorithm for this purpose is \emph{Partially Observable Monte-Carlo Planning (POMCP)}~\cite{Silver2010} which uses a particle filter to represent the belief and a Monte-Carlo Tree Search based strategy to compute the policy online. Recently, some techniques have been proposed to introduce prior knowledge in this algorithm \cite{Castellini2019,CastelliniAIRO2019}. The local representation of the policy made by this algorithm however hinders the interpretation and explanation of the policy itself.

Explainability~\cite{Gunning2019} is becoming a key feature of artificial intelligence systems since in several contexts humans need to understand why specific decisions are taken by the agent.
Specifically, explainable planning (XAIP)~\cite{Fox2017, Cashmore2019} focuses on explainability in planning methods. The presence of erroneous behaviors in these tools (due, for instance, to the wrong setup of internal parameters) may have a strong impact on autonomous cyber-physical and robotic systems that interact with humans, and detecting these errors in automatically generated policies is very hard in practice. For this reason, improving policy explanability is fundamental.

In this work, we propose a methodology for interpreting POMCP policies and detecting their unexpected decisions.
Using this approach, experts provide qualitative information on system behaviors (e.g., ``the robot should move fast if it is highly confident that the path is not cluttered'') and the proposed methodology supplies quantitative details of these statements based on evidence observed in traces (e.g., the approach says that the robot usually moves fast if the probability to be in a cluttered segment is lower than 1\%).

Experts are however also interested in identifying states in which the planner does not respect their assumptions. A possible question, in this case, is ``Is there a state in which the robot moves at high speed even if it is likely that the environment is cluttered?''.
To answer this kind of question, our approach allows expressing partially defined assumptions employing logical formulas, called \emph{rule templates}.
Parameters of rule templates are then computed from traces by a Satisfiability Modulo Theory (SMT) solver.

The approach we propose formalizes the parameter computation task as a \emph{MAX-SAT} problem which allows to express complex logical formulas and to compute optimal assignments when the template is not fully satisfiable (which happens in the majority of cases in real policy analysis).
A second key feature of the approach concerns the identification of decisions that violate trained rules.
Decision boundaries over beliefs generated by the proposed method have good interpretability and can be iteratively improved by focusing on specific questions that arise in the analysis of the policy.
This iterative process allows to locally explore policy behaviors.

Finally, since the proposed method quantifies the divergence between rule decision boundaries and decisions that do not satisfy the rules, it also identifies decisions that violate expert assumptions.
To empirically evaluate this feature, we inject some errors into POMCP by wrongly setting one of its parameters, and we show that our methodology can outperform standard anomaly detection methods identifying policy decisions that violate expert assumptions.
This performance improvement is achieved by exploiting the capability of the proposed method to include prior knowledge of the system under investigation.

The contribution of this paper to the state-of-the-art is threefold:
\begin{itemize}
    \item We propose a novel approach based on SMT for analyzing properties of POMCP traces by means of logical rules specified by human experts, variables are then instantiated by a MAX-SMT solver.
    \item We propose a method for identifying unexpected decisions using logical rules learned by the MAX-SMT solver. 
    \item We empirically evaluate the performance of the proposed method on two case studies, namely, the Tiger benchmark domain and a problem of velocity regulation for mobile robots.
\end{itemize}

\section{Related Works} \label{sec:related_works}
We have identified two main research topics with relationships with our method and goals, namely,  policy verification and explainable planning. Formal logic is strongly employed in verification of machine learning and reinforcement learning algorithms.

\paragraph{Policy verification} In recent years, SMT-based approaches have been developed to verify the safety of neural networks~\cite{Huang2017, Katz2017, Bunel18}. These methodologies encode the neural network into SMT formulas 
and check if safety properties hold on these formulas, or they provide counterexamples for properties that are not satisfied. 
To the best of our knowledge, there is no equivalent approach to verify that a specific property holds on policies generated by POMDPs, and in particular by POMCP.
A possibility to use SMT-based approaches to verify POMCP policies is to encode the POMDP problem in one of the logic-based frameworks presented in~\cite{Cashmore2016, Norman2017, Wang2018, Bastani2018}, where property guarantees can be formally proved. However,
these frameworks use SMT-solvers to build a policy that satisfies predefined properties while we use a MAX-SMT representation of the problem with a different goal, namely 
to evaluate if the policy satisfies expert assumptions. 
Namely, we aim at enhancing policy explainability without altering the policy itself.

A work in which verification is achieved by exploiting a simplified representation of the problem provided by an expert is~\cite{Zhu2019}. 
It describes a method for verifying properties related to the safety of fully observable systems modeled by Markov Decision Processes (MDPs). The approach works on a pre-trained neural network representing a black-box policy. It uses a linear formula summarizing the policy behavior to allow using off-the-shelf verification tools. This differs from our work for two reasons: first, we work on partially observable environments, and our logical formulas work on beliefs instead of states; second, in \cite{Zhu2019} the formula is used as an input to the verification tool while we use it to interact with humans and improve policy explainability.

\paragraph{Explainable planning} \emph{Explainable Artificial Intelligence (XAI)}~\cite{Gunning2019} is a rapidly growing research field focusing on human interpretability and understanding of artificial intelligence (AI) systems. In particular Explainable planning (XAIP)~\cite{Cashmore2019, Fox2017, Langley17, Sule2019} aims at investigating planning tools that come with justifications for the decisions they make. In our work, we use the high-level insight provided by the user to build an explanation of the policy in use.
A particularly interesting kind of questions analyzed in XAIP are known as \emph{contrastive question}~\cite{Fox2017}. They are used to structure the interaction between humans and the AI systems to be explained. In these questions, the expert asks the agent question as ``Why have you made this decision instead of this other one, that I believe could be a better option?''
and the system answers motivating its choice instead of the alternative one.
These questions are, however, very difficult to answer in online frameworks as POMCP~\cite{Castellini2020} because the information required to build the answer may not be available to the agent at run time. We, therefore, do not use contrastive questions but ground the interaction between human and planner on logical formulas that are framed by the expert, using her/his insight, and then instantiated by the SMT solver according to the observed behavior of the system.
The identification of decisions that violate user's expectation allows then to generate an iterative process in which the expert, that can refine the rule interactively, acquires new understanding about the policy. In \cite{CastelliniAIRO2020} a preliminary study on the use of MAX-SMT for explaining POMCP policies was proposed.

\section{Methods} \label{sec:method}
We provide full description of the proposed method using a running example to show direct application of the main concepts. 


\subsection{Method overview}
\begin{figure}[!ht]
\centering
\includegraphics[width=1.0\columnwidth]{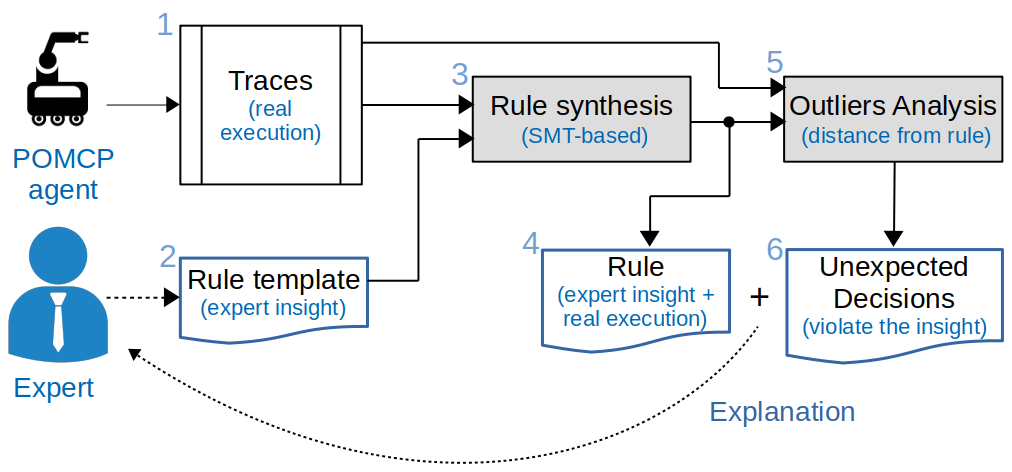} 
\caption{Methodology overview.} \label{fig:method_overview}
\Description{Summary of XPOMCP. A robot produces a trace (box 1), and an expert writes a rule template (box 2). An arrow from box 1 and an arrow from box 2 enter into the rule synthesis block (box 3). The result of rule synthesis is an arrow that splits in two: one enters in a box for the final rule (box 4) and one in a box for the outliers analysis algorithm (box 5). There is also an arrow from box 1 to box 5 to show that the outlier analysis uses the trace. The result of the outlier analysis is a set of unexpected decisions (box 6), shown with an arrow that points from box 5 to box 6. Finally, there is a dotted line that points back from the results (boxes 4 and 6) to the expert, to indicate that the process is interactive.}
\end{figure}

The methodology proposed in this work, called \emph{XPOMCP} in the following, is summarized in Figure~\ref{fig:method_overview}.
It leverages the expressiveness of logical formulas to represent specific properties of the investigated policy.
As a first step, a logical formula with free variables is defined (see box 2 in Figure~\ref{fig:method_overview}) to describe a property of interest of the policy under investigation. This formula, called \emph{rule template}, defines a relationship between some properties of the belief (e.g., the probability to be in a specific state) and an action.
Free variables in the formula allow the expert to avoid quantifying the limits of this relationship. These limits are then determined by analyzing a set of observed traces (see box 1). For instance, a template saying ``Do this when the probability of avoiding collisions is at least $\freevar{x}$'', with $\freevar{x}$ free variable, is transformed into ``Do this when the probability of avoiding collisions is at least $0.85$''.
By defining a rule template the expert provides useful prior knowledge about the structure of the investigated property.
Hence, the rule template defines the \emph{question} asked by the expert.
The \emph{answer} to this question is provided by the SMT solver (see box 3), which computes optimal values for the free variables in order to make the formula explain as many actions as possible in the observed traces.

The rule (see box 4) provides a human-readable local representation of the policy function that incorporates the prior knowledge specified by the expert, and it allows to split trace steps into two classes, namely, those satisfying the rule and those not satisfying it.
The approach, therefore, allows identifying unexpected decisions (see box 6), related to actions that violate the logical rule (i.e., that do not verify the expert's assumption).
The quantification of the violation, i.e., the distance between the rule boundary and the violation, also supports the analysis because it provides an explicit way to explain the violations themselves, which could even be completely unexpected due to expert imprecise knowledge or policy errors.

\subsection{Running example: velocity regulation in mobile navigation} \label{subsec:obstacle_avoidance}
\begin{figure}[!ht]
\centering
\includegraphics[width=0.95\columnwidth]{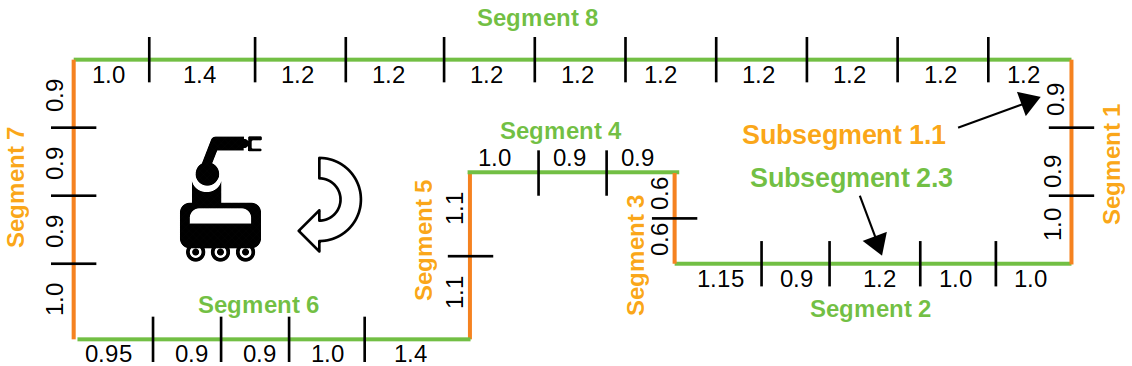} 
\textbf{(a)}
\begin{minipage}{.50\linewidth}
\centering
\small
\begin{tabular}{ccc}
    \hline
    $f$ & \hspace{0.2cm} & $p(o=1 \ | \ f)$ \\
    \midrule
    0 & & 0.0\\
    1 & & 0.5\\
    2 & & 1.0\\
    \hline
     & & \textbf{(b)\phantom{111}}\\
\end{tabular}
\end{minipage}%
\begin{minipage}{.50\linewidth}
\small
\centering
\begin{tabular}{rrr}
    \hline
    $f$ & $a$  & $p(c=1 \ | \ f,a)$ \\
    \midrule
    0 & 0 & 0.0\\
    0 & 1 & 0.0\\
    0 & 2 & 0.028\\
    1 & 0 & 0.0\\
    1 & 1 & 0.056\\
    1 & 2 & 0.11\\
    2 & 0 & 0.0\\
    2 & 1 & 0.14\\
    2 & 2 & 0.25\\
    \hline
    & & \textbf{(c)\phantom{111111}}\\
    \end{tabular}
\end{minipage}
\caption{Main elements of the POMDP model for the velocity regulation problem. (a) Path map. The map presents the length (in meters) for each subsegment. (b) Occupancy model $p(o \ | \ f)$: probability of observing a subsegment occupancy given segment difficulty. (c) Collision model $p(c\ | \ f,a)$: collision probability given segment difficulty and action.} \label{fig:lab_ice}
\Description{Map of the laboratory. It shows a robot that moves around the map. Themap has eight segments of different size. The number of subsegments in each segment is, in order: 3, 5, 2, 3, 2, 5, 4, 11. The size (in meter) of each subsegment is, in order: 0.9, 0.9, 1.0, 1.0, 1.0, 1.2, 0.9, 1.15, 0.6, 0.6, 0.9, 0.9, 1.0, 1.1, 1.1, 1.4, 1.0, 0.9, 0.9, 0.95, 1.0, 0.9, 0.9, 0.9, 1.0, 1.4, 1.2, 1.2, 1.2, 1.2, 1.2, 1.2, 1.2, 1.2, 1.2}
\end{figure}

We present a problem of velocity regulation in robotic platforms as a case study to show how XPOMCP works.
The same problem is used also in Section \ref{sec:results} to evaluate the performance of our method.
A robot travels on a pre-specified path divided into eight \emph{segments} which are in turn divided into \emph{subsegments} of different sizes, as shown in Figure~\ref{fig:lab_ice}.
Each segment has a (hidden) difficulty value among \emph{clear} ($f = 0$, where $f$ is used to identify the difficulty), \emph{lightly obstructed} ($f = 1$) or \emph{heavily obstructed} ($f = 2$). 
All the subsegments in a segment share the same difficulty value, hence the hidden state-space has $3^8$ states.
The goal of the robot is to travel on this path as fast as possible while avoiding collisions.
In each subsegment, the robot must decide a \emph{speed level} $a$ (i.e., action).
We consider three different speed levels, namely 0 (slow), 1 (medium speed), and 2 (fast).
The reward received for traversing a subsegment is equal to the length of the subsegment multiplied by $1+a$, where $a$ is the speed of the agent, namely the action that it selects.
The higher the speed, the higher the reward, but a higher speed suffers a greater risk of collision (see the collision probability table $p(c=1 \ | \ f,a)$ in Figure \ref{fig:lab_ice}.c).
The real difficulty of each segment is unknown to the robot (i.e., hidden part of the state), but in each subsegment, the robot receives an observation, which is $0$ (no obstacles) or $1$ (obstacles) with a probability depending on segment difficulty (see Figure \ref{fig:lab_ice}.b). 
The state of the problem contains a hidden variable (i.e., the difficulty of each segment), and three observable variables (current segment, subsegment, and time elapsed since the beginning).

We are interested in a rule describing when the robot travels at maximum speed (i.e., $a=2$).
We expect that the robot should move at that speed only if it is confident enough to be in an easy-to-navigate segment, but this level of confidence varies slightly from segment to segment (due to the length of the segments, the elapsed times, or the relative difficulty of the current segment in comparison to the others).
To obtain a rule that is compact but informative, we want the rule to be a local approximation of the behavior of the robot, thus we only focus on the current segment without considering the path as a whole when we write this rule.
The task of the proposed method is to find the actual bounds on the probability distribution (i.e., belief) that the POMCP algorithm uses to make its decisions and to highlight the (unexpected) decisions that do not comply with this representation.

\subsection{Partially Observable Monte-Carlo Planning}
A Partially Observable Markov Decision Process (POMDP)~\cite{Kaelbling98} is a tuple $(S, A, O, T, Z, R, \gamma)$,
where $S$ is a set of partially observable \emph{states},
$A$ is a set of \emph{actions},
$Z$ is a finite set of \emph{observations},
$T$:~$S\times A \rightarrow \Pi(S)$ is the \textit{state-transition model}, with $\Pi(S)$ probability distribution over states,
$O$:~$S\times A \rightarrow \Pi(Z)$ is the \textit{observation model},
$R$:~$S\times A \rightarrow \mathbb{R}$ is the \textit{reward function} and $\gamma \in [0,1]$ is a \textit{discount factor}.
An agent must maximize the \emph{discounted return} $E[\sum_{t=0}^{\infty} \gamma^t R(s_t,a_t)]$. 
A probability distribution over states, called \emph{belief}, is used to represent the partial observability of the true state.
To solve a POMDP it is required to find a \emph{policy}, namely a function $\pi$:~$B \rightarrow A$ that maps beliefs $B$ into actions.

We use \emph{Partially Observable Monte-Carlo Planning (POMCP)}~\cite{Silver2010} to solve POMDPs. POMCP is an \emph{online} algorithm that solves POMDPs by using Monte-Carlo techniques. The strength of POMCP is that it does not require an explicit definition of the transition model, observation model, and reward. Instead, it uses a black-box to simulate the environment.
POMCP uses a \emph{Monte-Carlo Tree Search} (MCTS) at each time-step to explore the belief space and select the best action. \emph{Upper Confidence Bound for Trees (UCT)}~\cite{Kocsis2006} is used as a search strategy to select the subtrees to explore and balance exploration and exploitation.
The belief is implemented as a \emph{particle filter}, which is a sampling over the possible states that is updated at every step.
At each time-step a particle is selected from the filter, each particle represents a state.
This state is used as an initial point to perform a simulation in the Monte-Carlo tree.
Each simulation is a sequence of action-observation pairs and it collects a discounted return, and for each action, we can compute the expected reward that can be achieved.
The particle filter is updated after receiving an observation.
If required, new particles can be generated from the current state through a process of \emph{particle reinvigoration}.

In the following, we call \emph{trace} a set of runs performed by POMCP on a specific problem. Each run is a set of \emph{steps}, and each step corresponds to an action performed by the agent having a belief and receiving an observation from the environment. In the velocity regulation problem,  we use 100 runs per trace.

\subsection{SMT and MAX-SMT}
The problem of reasoning on the satisfiability of formulas involving propositional logic and first-order theories is called  \emph{Satisfiability Modulo Theory} (SMT).
In XPOMCP, we use propositional logic and the theory of linear real arithmetic to encode the rules that describe the behavior of policies, and we use Z3~\cite{DeMoura2008} to solve the SMT problem.
We encode our formulas as a MAX-SMT problem, which has two kinds of clauses, namely, \emph{hard}, that must be satisfied, and \emph{soft} that can be satisfied.
A model of the MAX-SMT problem hence satisfies all the hard clauses and as many soft clauses as possible, and it is unsatisfiable only when hard clauses are unsatisfiable.
Our rules are intended to describe as many decisions as possible among those taken by the policy hence MAX-SMT provides a perfect formalism to encode this requirement.
The Z3 solver is used to solve the MAX-SMT problem~\cite{Bjorner2014}. Subsection~\ref{subsec:rulesynth} presents the details of this encoding.

The key ingredient for the MAX-SMT formulation are \emph{rules} and \emph{rule templates}.
A rule template represents the \emph{question} the expert wants to investigate.
It is a set of first-order logic formulas without quantifiers explaining some properties of the policy, and has the following form:
\begin{align}
    \begin{split}
        & \texttt{$r_1:$ select $a_1$ when (${\bigvee}_{i^1}$ subformula$_{i^1}$);} \\
        & \dots\\
        & \texttt{$r_n:$ select $a_n$ when (${\bigvee}_{i^n}$ subformula$_{i^n}$);} \\
        & \texttt{{[ where~${\bigwedge}_j$ (requirements$_j$); ]}}
    \end{split}
\end{align}
where $r_1,\dots,r_n$ are \emph{action rule templates}.
A \emph{subformula} is defined as ${\bigwedge}_k p_{s} \approx \freevar{x}_k$, where $p_s$ is the probability of state $s$, symbol $\approx \in \{<, >, \ge, \le\}$, and $\freevar{x}_k$ is a free variable that is automatically instantiated by the SMT solver analyzing the traces (when the problem is satisfiable). In general, bold letters with an overline (e.g., $\freevar{x}, \freevar{y}$) are used to identify free variable while italic letters (e.g., $p, a_i$) are used for fixed values read from the trace.
The \texttt{where} statement can be used to specify an optional set of hard requirements that can take different forms, such a the definition of a minimum value (e.g., $\freevar{x}_0 \ge 0.9$) or a relation (e.g., $\freevar{x}_2 = \freevar{x}_3$). These are used to define prior knowledge on the domain which is used by the rule synthesis algorithm to compute optimal parameter values (e.g., equality between two free-variables belonging to different rules can be used to encode the idea that two rules are symmetrical).

For instance, in our running example the actions $a_0, a_1$ and $a_2$ represent speeds (i.e., low, medium, high).
Each step $t$ contains the partially observable state 
$(\mathit{segment}^t, \mathit{subsegment}^t, \mathit{difficulty}^t)$ and the selected action $a^t$.
Since \emph{difficulty} is a probability distribution on $3^8=6561$ states we do not use this value directly.
For the sake of brevity, we introduce the \texttt{diff} function which takes a distribution on the possible difficulties \texttt{distr}, a segment \texttt{seg}, and a required difficulty value \texttt{d} as input and returns the probability that segment \texttt{s} has difficulty \texttt{d} in the distribution \texttt{distr}.
We can now write the rule template:
\begin{gather} \label{formula:complex_template}
    \begin{split}
    & \texttt{$r_2:$ select $a_2$ when~} p_0 \ge \freevar{x}_1 \lor p_2 \le \freevar{x}_2; \\
    & \texttt{where}~ \freevar{x}_1 \ge 0.9~\land~p_0 = \texttt{diff(distr, seg, 0)}~\land \\
    & \hspace{0.9cm} p_2 = \texttt{diff(distr, seg, 2)} \\
    \end{split}
\end{gather}
The first literal of $r_2$ specifies that we select action $a_2$ if the probability to be in a segment with low difficulty is greater than a certain threshold, where with $\freevar{x}_1 \ge 0.9$ in the requirement we declare that this threshold must be at least $0.9$ (an information that we expect to be true),
while the second is an upper bound on the belief that the current segment is hard (i.e., $p_2 \leq \freevar{x}_2$).
From now on, we always assume $p_{d} = \texttt{diff}(\mathit{difficulty}, \mathit{segment}, d)$ for $d \in \{0, 1, 2\}$ in the context of the velocity regulation problem.
To encode Equation~(\ref{formula:complex_template}), for each step $t$ in the trace we add the clauses:
\begin{itemize}
    \item $p^t_0 = \texttt{diff}(\mathit{difficulty}^t, \mathit{segment}^t, 0)$,
    \item $p^t_2 = \texttt{diff}(\mathit{difficulty}^t, \mathit{segment}^t, 2)$,
    \item if the robot performs action $a_2$ (moving fast), then the formula $(p^t_0 \le \freevar{x}_1 \lor p^t_2 \ge \freevar{x}_2)$ is added to the problem,
    \item if the robot performs a different action (i.e., $a_0$ or $a_1$) then the formula $\lnot (p^t_0 \le \freevar{x}_1 \lor p^t_2 \ge \freevar{x}_2)$ is added.
\end{itemize}
Finally, we add the constraints:
\begin{itemize}
    \item $(\freevar{x}_1 \ge 0.0 \land \freevar{x}_1 \le 1.0) \land (\freevar{x}_2 \ge 0.0 \land \freevar{x}_2 \le 1.0)$ to ensure that $\freevar{x}_1$ and $\freevar{x}_2$ are probabilities, 
    \item $\freevar{x}_1 \ge 0.9$ to force the hard constraint. 
\end{itemize}
A \emph{learned rule} is a rule template with all free variables instantiated (e.g., $\freevar{x}_1, \freevar{x}_2$).
For a rule to properly describe a trace generated by a policy, all steps in the trace should satisfy the rule (i.e., the action defined in the rule should be taken in a step iff the belief satisfies the rule conditions).
This is however almost impossible in real traces because the policy is usually a complex formula.
For this reason, we implemented a soft mechanism to check clause satisfiability, as described in subsection~\ref{subsec:rulesynth}.
In our example, from rule template~(\ref{formula:complex_template}) and a trace we obtain the learned rule:
\begin{gather*}
\texttt{$r_2:$ select $a_2$ when~} p_0 \ge 0.945 \lor p_2 \le 0.07;
\end{gather*}
while an example of output provided by XPOMCP when a rule is violated is:
\begin{align*}
& \texttt{Violation in run 2, step 4:}\\
& \texttt{- Selected action: $a_2$}\\
& \texttt{- Belief: $p_0=0.38, p_1 = 0.31, p_2 = 0.31$.}
\end{align*}

\subsection{Rule Synthesis} \label{subsec:rulesynth}
Rule synthesis is performed by Algorithm~\ref{algo:rule_synthesis} that takes as input a trace \emph{ex} generated by POMCP and a rule template \emph{r}. The output is the rule \emph{r} with all free variables instantiated to satisfy as many steps of \emph{ex} as possible. The \emph{solver} is a Z3 instance used to find a model for the formulas. 
\begin{algorithm}[ht]
    \KwData{a trace generated by POMCP $ex$\\\hspace{0.965cm}a rule template $r$}
    \KwResult{an instantiation of $r$
    }
    $solver \leftarrow$ probability constraints for thresholds in $r$\; \label{line:probability_axioms}
    \ForEach{action rule $r_a$ with $a\in A$} {\label{line:begin_dummy}
        \ForEach{step $t$ in $ex$} {
            build new dummy literal $l_{a,t}$\; \label{line:init_literal}
            $cost \leftarrow cost \cup l_{a,t}$\; \label{line:init_cost}
            compute $p_0^t, \dots, p_n^t$ from \emph{t.particles}\; \label{line:particles}
            $r_{a,t} \leftarrow $ instantiate rule $r_a$ using $p_0^t, \dots, p_n^t$\;\label{line:instatiation}
            \If{$t.action \neq a$} { 
                $r_{a,t} \leftarrow \lnot(r_{a,t})$\;\label{line:negation}
            }
            $solver.add(l_{a,t} \lor r_{a,t})$\;
             \label{line:end_if_action}
        } 
    } \label{line:end_dummy}
    
    $solver$.minimize($cost$)\; \label{line:minimize_cost}
    $goodness \leftarrow 1 - distance\_to\_observed\_boundary$\; \label{line:interval}
    $model \leftarrow$ $solver$.maximize($goodness$)\; \label{line:maximize_rule}
    \Return{$model$} \label{line:return_rule}
    \caption{RuleSynthesis} \label{algo:rule_synthesis}
\end{algorithm}

The \emph{solver} is first initialized and hard constraints are added in line~\ref{line:probability_axioms} to force all parameters in the template to satisfy the probability constraint (i.e., to have value in range $[0, 1]$).
Then in the \emph{foreach} loop in lines \ref{line:begin_dummy}--\ref{line:end_dummy} the algorithm maximizes the number of steps satisfying the rule template $r$.
In particular, for each action rule $r_a$, where $a$ is an action, and for each step $t$ in the trace $ex$ the algorithm first generates a literal $l_{a,t}$ (line \ref{line:init_literal}) which is a dummy variable used by MAX-SMT to satisfy clauses that are not satisfiable by a free variable assignment.
This literal is then added to the \emph{cost} objective function (line \ref{line:init_cost}) which is a pseudo-boolean function collecting all literals.
This function essentially counts the number of fake assignments that correspond to unsatisfied clauses.
Afterwards, the belief state probabilities are collected from the particle filter (line \ref{line:particles}) and used to instatiate the action rule template $r_a$ (line \ref{line:instatiation}) by substituting their probability variables $p_i$ with observed belief probabilities.
This generates a new clause $r_{a,t}$ which represents the constraint for step $t$.
This constraint is considered in its negated form $\lnot(r_{a,t})$ if the step action $t.action$ is different from $a$ (line \ref{line:negation}) because the clause $r_{a,t}$ should not be true.

The set of logical formulas of the solver is then updated by adding the clause $l_{a,t} \lor r_{a,t}$.
In this way the added clause can be satisfied in two ways, namely, by finding an assignment of the free variables that makes the clause $r_{a,t}$ true (the expected behavior) or by assigning a true value to the literal $l_{a,t}$ (unexpected behavior).
The second kind of assignment however has a cost since the dummy variables have been introduced only to allow partial satisfiability of the rules.
In line \ref{line:minimize_cost}, in fact, the solver is asked to find an assignment of free variable which minimizes the cost function, which considers the number of dummy variables assigned to true.
This minimization is a typical MAX-SMT problem in which an assignment maximizing the number of satisfied clauses is found. 
Since there can be more than a single assignment of free variables that achieves the MAX-SMT goal, the last step of the synthesis algorithm (lines~\ref{line:interval}--\ref{line:maximize_rule}) concerns the identification of the assignment which is closer to the behavior observed in the trace.
This problem is solved by maximizing a goodness function which moves the free variables assignment as close as possible to the numbers observed in the trace, without altering the truth assignment of the dummy literals.
Notice that this problem concerns the optimization of real variables and it is solved by the linear arithmetic module.    

Theoretically, MAX-SMT is an NP-hard but in practice, Z3 can solve our instances in a reasonable time (as shown in Section~\ref{sec:results}).
The variable in the SMT problem are the free variables specified in the template (a constant number) and the dummy literals, that are linear on the size of the trace because the algorithm builds a clause for each step, and each clause introduces a new dummy literal.

\subsection{Identification of unexpected decisions} \label{sec:identify_outlier}
A key element of XPOMCP concerns the characterization of steps that fail to satisfy the rule.
They can be seen as \emph{unexpected decisions}, namely, exceptions to the general rule that the expert expects to be true.
They can provide useful information for policy interpretation.
We define two important classes of exceptions, namely, those related to the approximation made by the logical formula and those actually due to an unpredicted behavior (e.g., an error in the POMCP algorithm, or a decision that cannot be described with only local information).
We expect exceptions in the first class to fall quite close to the rule boundary, while exceptions in the second class to be more distant from the boundary.
In the following, we call the second kind of exceptions \emph{unexpected decisions} since their behavior is unexpected compared to the expert knowledge on the policy.

In this section, we provide a procedure for differentiating the first class of exceptions from the second one, to find as many unexpected decisions as possible.
The input of the procedure is a learned rule $r$, a set of steps (called \emph{steps}) that violate the rule, and a threshold $\tau \in [0,1]$.
The output of the algorithm is a set of steps related to unexpected decisions.
The procedure first randomly generates $w$ samples (i.e., beliefs) $\bar{b}_j, j=1,\ldots, w,$  that satisfy the rule. Then, for each belief $b_i$ in $steps$ a distance measure is computed between $b_i$ and all $\bar{b}_j, j=1,\ldots, w$.
The minimum distance $h_i$ is finally computed for each $b_i$ and compared to a threshold $\tau$.
If $h_i \geq \tau$ then $b_i$ is considered an outlier because its distance from the rule boundary is high. 

Since beliefs are discrete probability distributions, we use a specific distance measure dealing with such kinds of elements, namely, the \emph{discrete Hellinger distance ($H^2$)} \cite{Hellinger1909}.
This distance is defined as follow:
\begin{gather*}
    H^2(P, Q) = \frac{1}{\sqrt{2}} \sqrt{\sum_{i = 1}^{k} ( \sqrt{P_i} - \sqrt{Q_i} )^2}
\end{gather*}
where $P, Q$ are probability distributions and $k$ is the discrete number of states in $P$ and $Q$. An interesting property of $H^2$ is that it is bounded between $0$ and $1$, which is very useful to define a meaningful threshold $\tau$. In Section~\ref{sec:results} we discuss how we set this threshold for our experiments.


\section{Results} \label{sec:results}
This section provides experimental results on two case studies.
The capability of XPOMCP to identify unexpected decisions is compared to that of \emph{isolation forest}~\cite{Liu2008}, a state-of-the-art anomaly detection algorithm, showing that our method outperforms isolation forest in terms of F1-score and accuracy.

\subsection{Experimental Setting}
We implemented two problems, namely \emph{Tiger} and \emph{velocity regulation}, as black-box simulators in the original C++ version of POMCP~\cite{Silver2010}.
To generate \emph{traces}, we collect both particle distributions and actions selected at each step.
The \emph{RuleSynthesis} algorithm (i.e., Algorithm~\ref{algo:rule_synthesis}) and the procedure for identifying \emph{unexpected decisions} (see Section~\ref{sec:identify_outlier}) have been developed in Python. 
The Python binding of Z3~\cite{DeMoura2008} has been used to solve the SMT formulas.
Experiments have been performed on a notebook with Intel Core i7-6700HQ and 16GB RAM. The code is available at \texttt{https://github.com/GiuMaz/AAMAS2021}.

\paragraph{Error injection}
To quantify the capability of the proposed method to identify policy errors 
we modify the \emph{RewardRange} parameter (called $W$ in the following) in POMCP.
This parameter defines the maximum difference between the lowest and the highest possible reward, and it is used by UCT to balance exploration and exploitation.
If this value is lower than the correct one the algorithm could find a reward that exceeds the maximum expected value leading to a wrong state, namely, the agent believes to have identified the best possible action and it stops exploring new actions, even though the selected action is not the best one.
This is a creeping error that randomly affects the exploration-exploitation trade-off making POMCP incorrect in some situations.
We use this kind of error since parameter \emph{W} must be set by hand in POMCP  and it requires specific values that are not always easy to collect.

\paragraph{Exact solution}
We use the \emph{incremental pruning algorithm}~\cite{Cassandra97} implemented in~\cite{Bargiacchi2020} to compute an exact policy for \emph{Tiger}.
This is used as a ground-truth for evaluating the performance of our method in detecting wrong actions.
Unfortunately, we cannot compute the exact policy for the \emph{velocity regulation} problem since its dimension makes the computation intractable, however, we use this case study to evaluate the applicability of our method to larger problems.

\paragraph{Baseline method}
\emph{Isolation forest (IF)} \cite{Liu2008} is an anomaly detection algorithm that we use as a benchmark for evaluating the performance of our procedure in identifying unexpected decisions.
It assumes anomalies to be rare events and can be applied to a training set containing both nominal and anomalous samples, hence it is a good candidate for comparison with XPOMCP. 
We use the Python implementation of IF provided in \emph{scikit-learn}~\cite{Pedregosa2011} and consider each step of a trace (i.e., a pair \emph{belief, action}) as a sample (notice that the action is not used as a label).
The algorithm uses the \emph{contamination} parameter (i.e., the expected percentage of anomalies in the dataset) to set the threshold used to identify which points are anomalies.

\subsection{Results on Tiger}
\emph{Tiger} is a well-known problem~\cite{Kaelbling98} in which an agent has to chose which door to open among two doors, one hiding a treasure and the other hiding a tiger. Finding the treasure yields a reward of $+10$ while finding the tiger a reward of $-100$. The agent can also listen (by paying a small penalty of $-1$) to gain new information. Listening is however not accurate since there is a $0.15$ probability of hearing the tiger from the wrong door.
A successful policy should listen until enough information is collected about the position of the tiger and then open a door when the agent is reasonably certain to find the treasure behind it.
From the analysis of the observation model and reward function, it is however not immediate to define what ``reasonably certain'' means.
To investigate it, we create a rule template specifying a relationship between the confidence (in the belief) over the treasure position and the related opening action.
Then we learn the rule parameters from a set of runs (i.e., a trace) performed using POMCP.
Finally, by analyzing the trained rule we understand which is the minimum confidence required by the policy to open a door.
The correct value of \emph{W} is $110$ (the reward interval is $[-100,10]$).
For each value of $W$ in $\{110, 85, 65, 40\}$, we generate $50$ traces with $1000$ runs each, using different seeds for the pseudo-random algorithm in every trace.
For each run,  we use $2^{15}$ particles and a maximum of $10$ steps.
Lower values of $W$ produce a higher number of errors, as show in Table~\ref{tab:tiger_comparison} (see column \emph{\% errors}).

\paragraph{Rule synthesis}
To formalize the property that the agent has to gather enough confidence on the tiger position before opening a door we use the following rule template:
\begin{align}
    \begin{split}
    & \texttt{$r_L:$ select}~Listen~\texttt{when~} (p_{right} \le x_1 \land p_{left} \le x_2);\\
    & \texttt{$r_{OR}:$ select}~Open_{R}~\texttt{when~} p_{right} \ge x_3;\\
    & \texttt{$r_{OL}:$ select}~Open_{L}~\texttt{when~} p_{left} \ge x_4;\\
    & \texttt{where } (x_1 = x_2) \land (x_3 = x_4) \land (x_3 > 0.9);
    \end{split}
\end{align}
Action rule template $r_L$ describes when the agent should listen, while templates $r_{OR}$ and $r_{OL}$ describe when the agent should open the right and left door, respectively.
Some hard clauses are also added (in the bottom) to state that \emph{i)} the problem is expected to be symmetric (i.e. the thresholds used to decide when to listen and when to open are the same for both doors, namely  $x_1 = x_2$ and $x_3 = x_4$), \emph{ii)} a minimum confidence is expected to open the door (namely, $x_3>0.9$, hence the door should be opened only if the agent is at least 90\% sure to find the tiger behind it).

\paragraph{Performance evaluation regardless of threshold}
Both XPOMCP and IF use a threshold to identify anomalous points.
We use the \emph{Receiver Operating Characteristic (ROC) curve} and the \emph{precision/recall curve} of the two methods to compare the performance across thresholds.
The ROC curve considers the relationship between the true positive rate (tpr) and the false positive rate (fpr) at different thresholds.
We use the \emph{Area Under Curve} (AUC) as a performance measure.
Similarly, the precision/recall curve considers the relationship between precision and recall at different thresholds and we use the \emph{Average Precision} (AP) as a performance measure.
Performance are compared on traces generated using $W \in \{85, 65, 40\}$. We do not evaluate the methods in the case with $W=110$, it is error-free thus it is not possible to have any true positive (AUC and AP are~$0$).

\begin{figure}[t]
\includegraphics[width=1\linewidth]{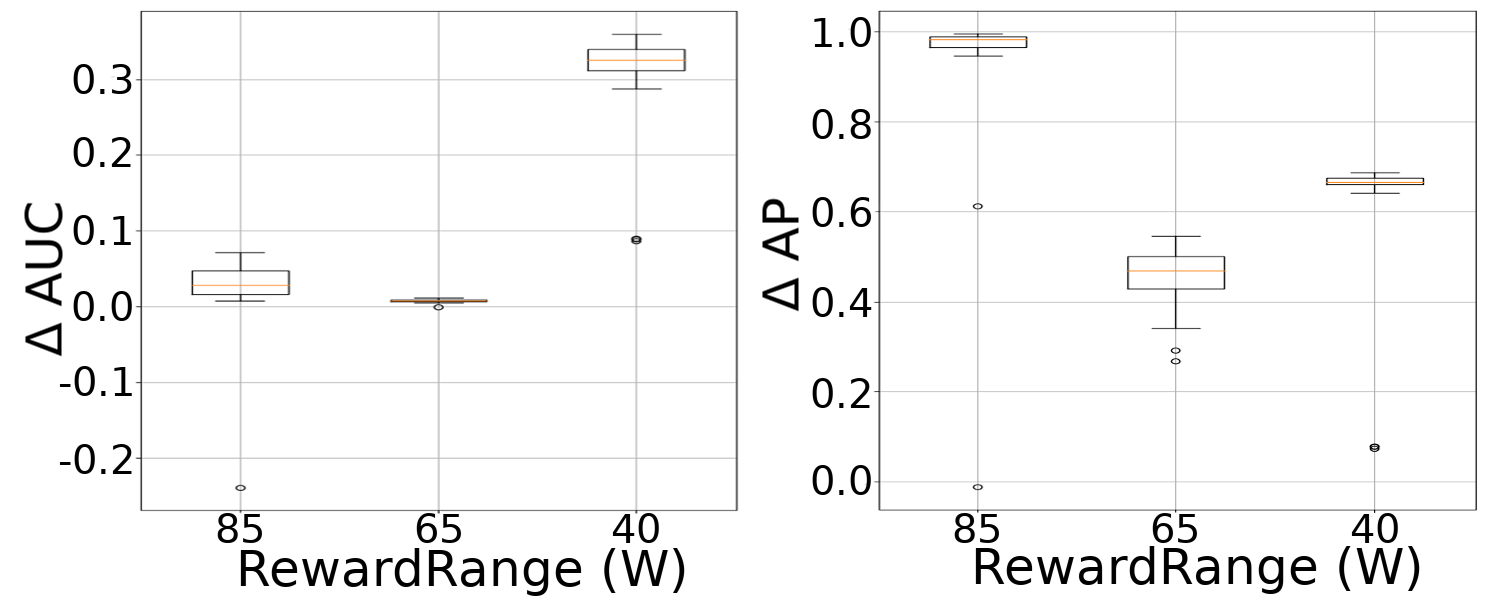}
\caption{Box-plots of AUC and AP (considering 100 different parameters) with different values of \emph{W}} \label{fig:ap_auc}
\Description{Two box-plots that show the difference in performance between XPOMCP and Isolation Forest. Each of the two consider three experiment (W = 85, 65, 40). The first box-plot shows the difference of AUC. The second one shows the difference of AP.}
\end{figure}

In Table~\ref{tab:tiger_comparison} we compare the average performance of the two methods.
We test XPOMCP with a uniform sampling of $100$ thresholds in the interval $[0, 0.5]$. Similarly, we use IF with $100$ different values for the contamination parameter uniformly distributed in the interval $[0, 0.5]$.
XPOMCP outperforms IF in nearly every instance in both AUC and AP.
For both AUC and AP, the difference is high in the case of \emph{W=40}.
This is because XPOMCP effectively exploits the information in the template to avoid being influenced by the number of errors.
IF performs poorly also in the case of \emph{W=85} because it exhibits a large number of false-positive that leads to very low precision and value of AP.
Finally, in the case of \emph{W=65}, the difference between the two methods is smaller.
In this data-set, both methods achieve their best performance. IF is more effective in identifying true error compared to the case of $W=85$, but it still generates more false-positive than XPOMCP.
Figure~\ref{fig:ap_auc} displays two boxplots that show how values $\Delta AUC = AUC_{XPOMCP} - AUC_{IF}$ and $\Delta AP = AP_{XPOMCP} - AP_{IF}$ vary in each trace.
Since a positive value in the box-plot means that XPOMCP outperforms IF the plot shows that our algorithm is consistently better than IF except for an outlier in the case $W=85$.

\begin{table}[t]
\caption{Quantitative performance comparison (AUC and AP) at different values of \emph{W}. Best results are bold} \label{tab:tiger_comparison}
\centering
{
\subcaption{XPOMCP}
\begin{tabular}{rrrr}
\hline
\emph{W}        & \textbf{\% errors}   & \textbf{AUC}                & \textbf{AP}                \\
\midrule
$110$           & $0.0 (\pm 0.0)$      &                --           &                 --         \\
$85$            & 0.0004($\pm$0.0003)  & \textbf{0.993}($\pm$0.041)  & \textbf{0.986}($\pm$0.082) \\
$65$            & 0.0203($\pm$0.0021)  & \textbf{0.999}($\pm$0.001)  & \textbf{0.999}($\pm$0.002) \\
$40$            & 0.2374($\pm$0.0072)  & \textbf{0.995}($\pm$0.034)  & \textbf{0.987}($\pm$0.084) \\
\hline
\end{tabular}
}

\bigskip
{
\subcaption{Isolation Forest}
\begin{tabular}{rrrr}
\hline
\emph{W}        & \textbf{\% errors}    & \textbf{AUC}       & \textbf{AP}         \\
\midrule
$110$           & $0.0 (\pm 0.0)$       &                 -- &                 --  \\
$85$            & $0.0004 (\pm 0.0003)$ & $0.964(\pm 0.024)$ & $0.057(\pm 0.1076)$ \\
$65$            & $0.0203 (\pm 0.0021)$ & $0.992(\pm 0.001)$ & $0.539(\pm 0.0520)$ \\
$40$            & $0.2374 (\pm 0.0072)$ & $0.675(\pm 0.020)$ & $0.333(\pm 0.0153)$ \\
\hline
\end{tabular}
}
\end{table}

\begin{figure}[t]
\includegraphics[width=1.0\linewidth]{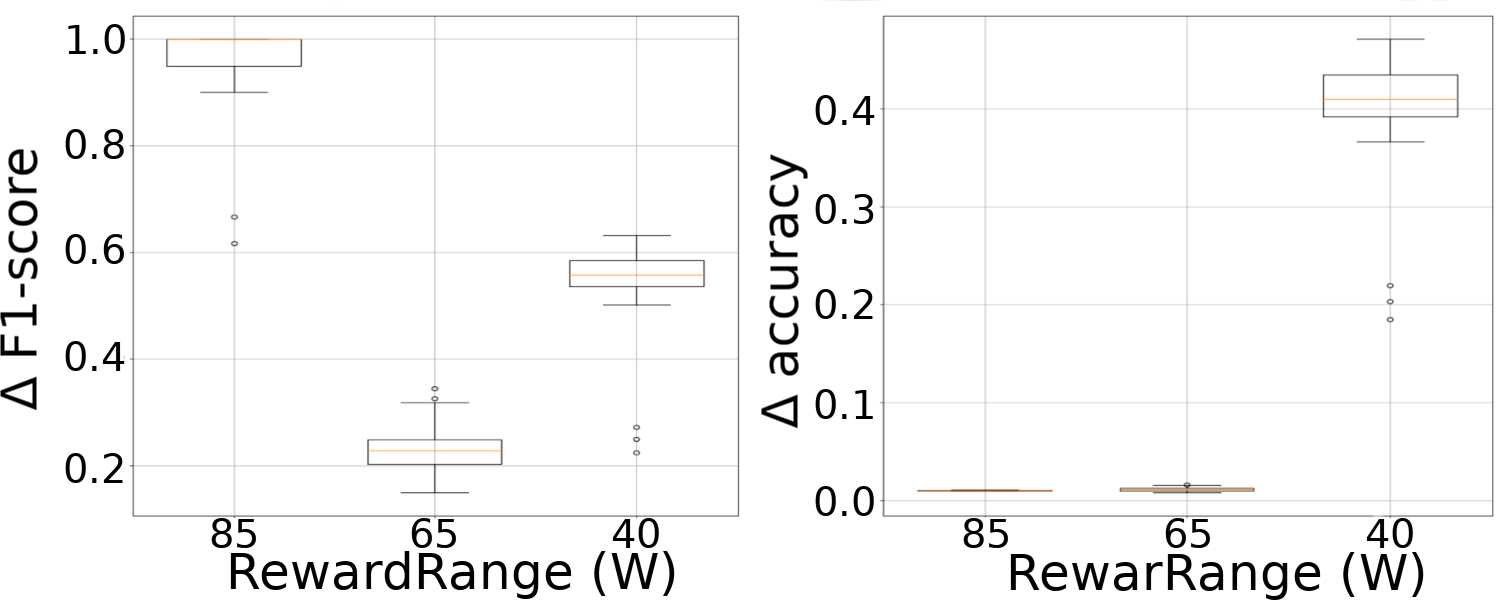}
\caption{Box-plots of $\Delta$ F1-score and $\Delta$ accuracy using the optimal thresholds with different values of \emph{W}} \label{fig:f1_acc}
\Description{Two box-plots that show the difference in performance between XPOMCP and Isolation Forest with an optimal threshold. Each of the two consider three experiment (W = 85, 65, 40). The first box-plot shows the difference of F1-score. The second one shows the difference of accuracy.}
\end{figure}

\paragraph{Performance evaluation with optimal parameters}
To provide further details on the performance of XPOMCP, here we show its performance with optimal threshold $\tau$ (see Section \ref{sec:identify_outlier}). To compute the value of $\tau$ we performed cross-validation by training XPOMCP on 5 traces and testing it on 45 traces. F1-score about the identification of unexpected decisions was computed on the test set using $100$ threshold values uniformly distributed in $[0, 0.5]$, and the test with the best F1-score was selected.
The results of this test are presented in Table~\ref{tab:tiger_optimal_threshold}.a.
Column \emph{threshold} contains values of threshold used and columns \emph{accuracy} and \emph{F1} show the related performance values on the test set, and \emph{time} shows the average elapsed time (in second).
We used the same procedure to tune the \emph{contamination} parameter of IF (Table~\ref{tab:tiger_optimal_threshold}.b).
Figure~\ref{fig:f1_acc} shows a comparison of the average F1-score and accuracy achieved by the two approaches in each test (the value in parenthesis presents the standard deviation).
This comparison shows that with optimal parameters XPOMCP always outperforms IF.
Both methods achieve high accuracy due to the high number of non-anomalous samples in the dataset (anomaly and non-anomaly classes are unbalanced) and several true negatives are computed by both methods.
However, the F1-score is very different. IF achieves a low score in this metric because it cannot identify some true positive and it generates much more false positives than XPOMCP.
In general, IF is faster than XPOMCP of an order of magnitude, but the performance of our methodology is acceptable since it takes POMCP an average of $158.2s$ to generate a \emph{Tiger} trace with 1000 runs and XPOMCP analyze it in less than $15s$.

\begin{table}[t]
\small
\caption{Quantitative performance comparison (F1 and accuracy) using optimal thresholds. Best results are bold} \label{tab:tiger_optimal_threshold}
\centering
{
\subcaption{XPOMCP}
\begin{tabular}{rcccc}
\hline
\emph{W} & \textbf{Threshold} & \textbf{F1}                & \textbf{Accuracy}           & \textbf{time (s)} \\
\midrule
$85$     & 0.061              & \textbf{0.979}($\pm$0.081) & \textbf{0.999}($\pm$0.0001) & 14.30($\pm$0.50) \\
$65$     & 0.064              & \textbf{0.999}($\pm$0.002) & \textbf{0.999}($\pm$0.0001) & 14.75($\pm$0.80) \\
$40$     & 0.045              & \textbf{0.980}($\pm$0.072) & \textbf{0.987}($\pm$0.049)  & 12.78($\pm$0.83) \\
\hline
\end{tabular}
}

\bigskip
{
\subcaption{Isolation Forest}
\begin{tabular}{rcccc}
\hline
\emph{W} & \textbf{Contamination} & \textbf{F1}  & \textbf{Accuracy} & \textbf{time (s)} \\
\midrule
$85$     & 0.01                   & 0.020($\pm$0.033) & 0.990($\pm$0.001)  & \textbf{0.72}($\pm$0.013) \\
$65$     & 0.03                   & 0.771($\pm$0.044) & 0.988($\pm$0.001)  & \textbf{0.71}($\pm$0.010) \\
$40$     & 0.5                    & 0.437($\pm$0.035) & 0.585($\pm$0.026)  & \textbf{0.64}($\pm$0.037) \\
\hline
\end{tabular}
}
\end{table}

\paragraph{Analysis of a specific trace}
To complete our analysis on \emph{Tiger}, we show the rule generated by XPOMCP on the analysis of a specific trace generated by POMCP using a wrong value of $W$ (i.e., $W=40$).
The rule generated by the MAX-SMT solver from this trace is:
\begin{align}
    \begin{split} \label{rule:correct}
     & \texttt{$r_{L}:$ select}~Listen~\texttt{when~} (p_{right} \le 0.847 \land p_{left} \le 0.847);\\
     & \texttt{$r_{OR}:$ select}~Open_{R}~\texttt{when~} p_{right} \ge 0.966;\\
     & \texttt{$r_{OL}:$ select}~Open_{L}~\texttt{when~} p_{left} \ge 0.966;
    \end{split}
\end{align}
It is a compact summary of the policy that highlights the important details in a structured way.
There is a gap between the value of rule $r_L$ (i.e., $0.8638$) and that of rules $r_{OR}, r_{OL}$ (i.e., 0.9644).
This is because the trace does not contain any belief in this gap and XPOMCP cannot build a rule to describe how to act in these beliefs.
An in depth analysis of this case study is performed in the Supplementary material.

\subsection{Results on the velocity regulation problem}
To evaluate the performance of XPOMCP on the \emph{velocity regulation} problem, we inspect by-hand each decision that is marked as unexpected, making a detailed analysis of traces (we recall that the exact policy cannot be computed for this problem because the state space is too long).
The template used to describe when the robot must move at high speed is as follow:
\begin{align*}
    & \texttt{$r_2:$ select action $S_{2}$ when~}  p_0 \ge \freevar{x}_1 \lor p_2 \le \freevar{x}_2~\lor \\
    &  \hspace{3.95cm}                           (p_0 \ge \freevar{x}_3 \land p_1 \ge \freevar{x}_4)\\
    & \texttt{where}~ \freevar{x}_1 \ge 0.9
\end{align*}
where $\freevar{x}_1,\freevar{x}_2,\freevar{x}_3,\freevar{x}_4$ are free variables and $p_0, p_1, p_2$ are defined as in Section~\ref{subsec:obstacle_avoidance}.
The first two constraints of the rule are identical to the running example, but we add a third constraint ($p_0 \ge \freevar{x}_3 \land p_1 \ge \freevar{x}_4$) that combines $p_0$ and $p_1$ to describe when the robot must move at high speed.
The correct value of $W$ for \emph{velocity regulation} is $103$ (i.e., the difference between moving at speed $1$ in a short segment and collide vs. going fast in a long subsegment without collisions, i.e., $0.6 \cdot 2 - 100, 1.4 \cdot 3$), but we set it to $90$ to generate some errors.
We run XPOMCP on a trace of $100$ runs and we obtain the rule:
\begin{align*}
    & \texttt{$r_2:$ select action $S_{2}$ when~}  p_0 \ge 0.910 \lor p_2 \le 0.013~\lor \\
    &  \hspace{3.95cm}                            (p_0 \ge 0.838 \land p_1 \ge 0.132)
\end{align*}
It takes XPOMCP $69.53s$ to analyze the trace.
This rule fails on $33$ out of $3500$ decisions, but only $4$ of them are marked by XPOMCP as unexpected using threhsold $\tau=0.1$, that we select empirically by analyzing the $H^2$ of unexpected decisions on \emph{velocity regulation} traces.
Table~\ref{tab:failed_steps_vr} shows some of the most notable steps that do not satisfy the rule (which are not necessarily unexpected decisions) in decreasing order of $H^2$.
Column \emph{\#} shows an identification number for the step, columns $p_0, p_1, p2$ show the belief of the step, column $H^2$ shows the Hellinger distance of the failed steps, and column \emph{unexp.} shows the outcome of the classification based on threshold $\tau=0.1$.
Steps~1 and 2 are unexpected behaviours since POMCP decided to move at high speed even if it had poor information on the difficulty of the segment ($p_0, p_1, p_2$ are close to a uniform distribution).
Steps number $3$ and $4$ are also unexpected. While they are closer to our rule, because $p_0$ is the dominant value in the belief, they are significantly distant from the boundary of the rule and the decision taken by POMCP.
Steps~5--33 cannot be satisfied due to the approximate nature of the rule but do not violate the expert indications.

\begin{figure}[t]
\centering
\includegraphics[width=1.0\columnwidth]{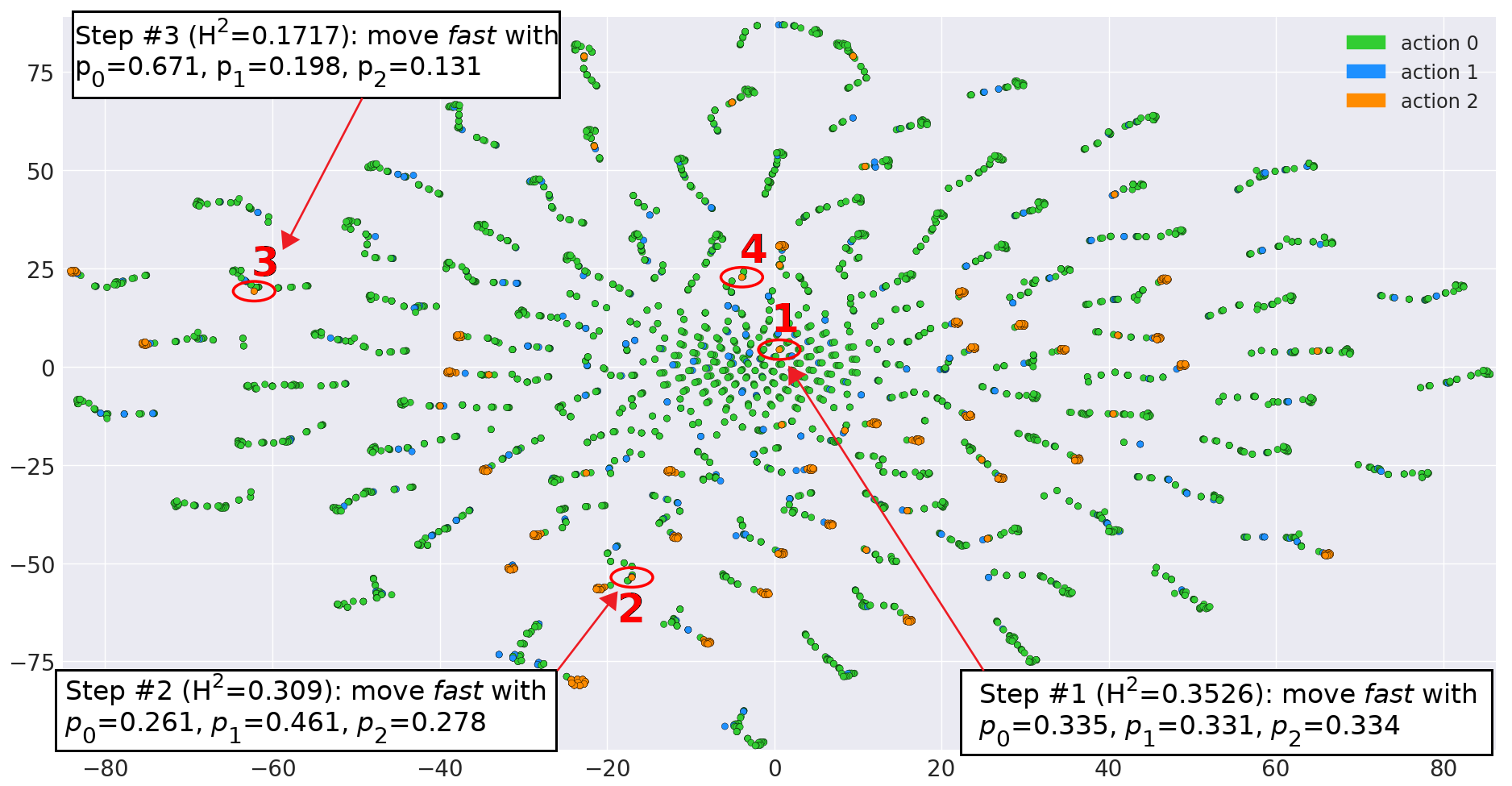} 
\caption{t-SNE of n-uples (belief, action) for the \emph{velocity regulation} with $W=90$. Our algorithm identify 4 anomalies, they are circled in red} \label{fig:tsne_obstacle_avoidance}
\end{figure}

To visualize the result of our approach we show a \emph{T-distributed Stochastic Neighbor Embedding (t-SNE) projection}~\cite{Hinton2008} in which the belief at each step is used to compute point coordinates and the action taken by POMCP is represented by different colors (see Figure~\ref{fig:tsne_obstacle_avoidance}).
In particular, green, blue, and orange points represent steps in the traces in which POMCP selected, respectively, a low speed, a medium speed ad a high speed.
While our rule generated by XPOMCP presents a clear and compact representation of the boundary on the belief that must be satisfied to select speed 2, there are no obvious separations between the points of the three speed values in the graph.
Most orange points are grouped in small clusters spread around the graph, but some isolated orange points are also present.
The steps that are classified as unexpected decisions are circled in red.
Points $1, 3$ are isolated and far from any small cluster of orange points while point $2$ and $4$ are close to one of the clusters.
Note that not all isolated points are marked as unexpected, XPOMCP identify the unexpected points not only by using their belief but also the insight provided by the expert.

\begin{table}[t]
\centering
\caption{Notable steps in velocity regulation ($W=90$)} \label{tab:failed_steps_vr}
\begin{tabular}{rccccc}
    \hline
    \textbf{\#} & $p_0$   & $p_1$   & $p_2$   & $H^2$    & \emph{unexp.} \\
    \midrule
    $1$         & $0.335$ & $0.331$ & $0.334$ & $0.3526$ & yes \\
    $2$         & $0.261$ & $0.461$ & $0.278$ & $0.3090$ & yes \\
    $3$         & $0.671$ & $0.198$ & $0.131$ & $0.1717$ & yes \\
    $4$         & $0.678$ & $0.228$ & $0.094$ & $0.1389$ & yes \\
    $5$         & $0.775$ & $0.196$ & $0.029$ & $0.0411$ & no  \\
    $6$         & $0.832$ & $0.127$ & $0.041$ & $0.0347$ & no  \\
    $32$        & $0.853$ & $0.126$ & $0.021$ & $0.0109$ & no  \\
    $33$        & $0.826$ & $0.160$ & $0.014$ & $0.0105$ & no  \\
    \hline
\end{tabular}
\end{table}

\balance
\section{Conclusions and future work} \label{sec:conclusion}
In this work, we present a methodology that combines high-level indications provided by a human expert with an automatic procedure that analyzes execution traces to synthesize key properties of a policy in the form of rules.
We exploit such rules to identify anomalous behavior, and we show that our methodology outperforms a state-of-the-art anomaly detection algorithm by exploiting the high-level indications.
This work paves the way towards several interesting research directions. Specifically, we aim at improving the expressiveness of the logical formulas used to formalize the indications of the expert (e.g., by employing temporal logic), and to develop an online interaction between POMCP and XPOMCP, where the rules are generated while POMCP is operating and not only on execution traces.


\bibliographystyle{ACM-Reference-Format} 
\bibliography{main} \label{sec:bib}


\end{document}


\maketitle 

\section{Analysis of a specific trace}
\begin{figure}[!ht]
\includegraphics[width=1.0\columnwidth]{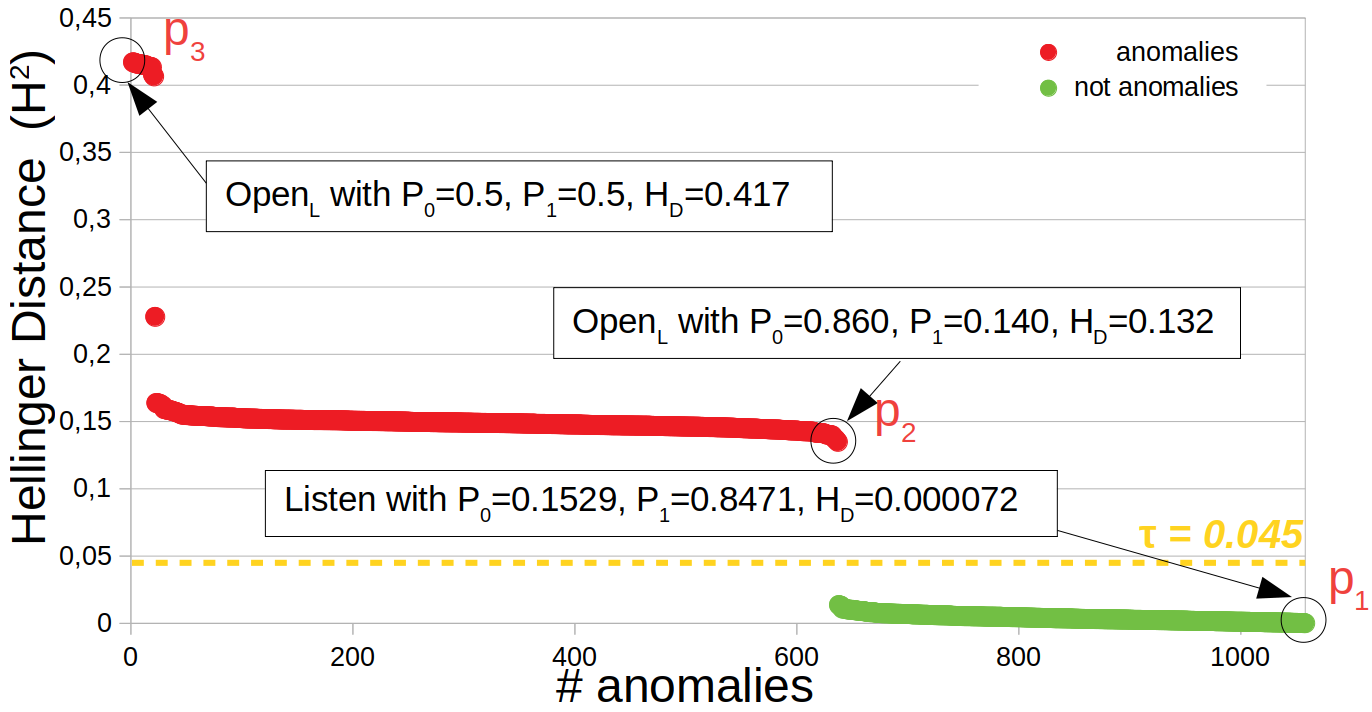} 
\caption{Values of Hellinger distance of steps that fail to satisfy the rule for \emph{Tiger}.} \label{fig:hellinger_plot}
\end{figure}

We show the results of XPOMCP on the analysis of a specific trace generated by POMCP using a wrong value of $W$ (i.e., $W=40$).
As shown in Table~\ref{tab:tiger_comparison}, POMCP selects several wrong actions with this value of $W$.
The rule generated by the MAX-SMT solver from this trace is:
\begin{align}
    \begin{split} \label{rule:correct}
     & \texttt{$r_{L}:$ select}~Listen~\texttt{when~} (p_{right} \le 0.847 \land p_{left} \le 0.847);\\
     & \texttt{$r_{OR}:$ select}~Open_{R}~\texttt{when~} p_{right} \ge 0.966;\\
     & \texttt{$r_{OL}:$ select}~Open_{L}~\texttt{when~} p_{left} \ge 0.966;
    \end{split}
\end{align}
It is a compact summary of the policy that highlights the important details in a structured way.
There is a gap between the value of the action rule $r_L$ (i.e., $0.847$) and that of the action rules $r_{OR}, r_{OL}$ (i.e., $0.966$).
This is because the trace does not contain any belief in this gap and XPOMCP cannot build a rule to describe how to act in this belief interval.
Among the total $2659$ trace steps, $1601$ satisfy the rule and $1058$ does not satisfy it.
For all steps not satisfying the rule, we computed their Hellinger distance using the procedure described in the article.
The $H^2$ distance values are displayed in descending order in Figure~\ref{fig:hellinger_plot}.

To classify unexpected decisions, we use the optimal threshold of $\tau=0.045$ (Section~4.2 of the main paper presents the process used to compute the optimal threshold).
We colored in green steps having a $H^2$ distance below the threshold and in red steps with $H^2$ above the threshold.
Notice that only the steps of the second group (which contains $637$ unsatisfiable steps in total) are considered unexpected decisions by our approach.
Three main levels of $H^2$ can be observed in the chart, one close to zero, one around $0.15$, and the third around $0.4$.
We have selected one step for each level (see points $p_1$, $p_2$ and $p_3$) and shown the related belief probabilities, action, and $H^2$ distance, to characterize the errors made by POMCP in those steps.
This analysis shows that in step $p_2$ POMCP decides to open the left door with a probability $0.860$ to have the treasure behind that door, while the rule says that a probability of at least $0.966$ is needed to select this action.
In step $p_3$ the error is even more evident since the agent opens the left door with a probability of only $0.5$ to find the treasure there.
On the other hand in step, $p_1$ the agent listens with a probability $0.8471$ which is slightly higher than the minimum value (i.e., $0.8470$) needed to perform that action.
Notice that this belief is not assigned to any action rule and it is very close to the boundary of the listen rule, hence it makes sense to assign this step to that rule and not to consider it as an exception.
In this case, we achieve an F1-score of $1.0$ and an accuracy of $1.0$ using the proposed approach, and it takes XPOMCP $12,509s$ to compute this solution.
This confirms the ability to detect unexpected decisions even when state-of-the-art anomaly detection methods have degraded performance (i.e., IF reaches an F1-score of $0.59$ in this test, but it only takes $0,791s$ to do so).

\begin{table}[h]
\centering
\caption{Quantitative performance comparison (AUC and AP) with uniform sampling of thresholds and different values of \emph{W}} \label{tab:tiger_comparison}
\begin{tabular}{rrrrrr}
\hline
\emph{W} & \textbf{\% errors}  & \textbf{XPOMCP AUC}        & \textbf{XPOMCP AP}         & \textbf{IF AUC}    & \textbf{IF AP}      \\
\midrule
$110$    & $0.0 (\pm 0.0)$     & --                         & --                         & --                 & --                  \\
$85$     & 0.0004($\pm$0.0003) & \textbf{0.993}($\pm$0.041) & \textbf{0.986}($\pm$0.082) & $0.964(\pm 0.024)$ & $0.057(\pm 0.1076)$ \\
$65$     & 0.0203($\pm$0.0021) & \textbf{0.999}($\pm$0.001) & \textbf{0.999}($\pm$0.002) & $0.992(\pm 0.001)$ & $0.539(\pm 0.0520)$ \\
$40$     & 0.2374($\pm$0.0072) & \textbf{0.995}($\pm$0.034) & \textbf{0.987}($\pm$0.084) & $0.675(\pm 0.020)$ & $0.333(\pm 0.0153)$ \\
\hline
\end{tabular}
\end{table}

\section{Larger t-SNE Figure}
\begin{figure}[!ht]
\includegraphics[width=1.0\columnwidth]{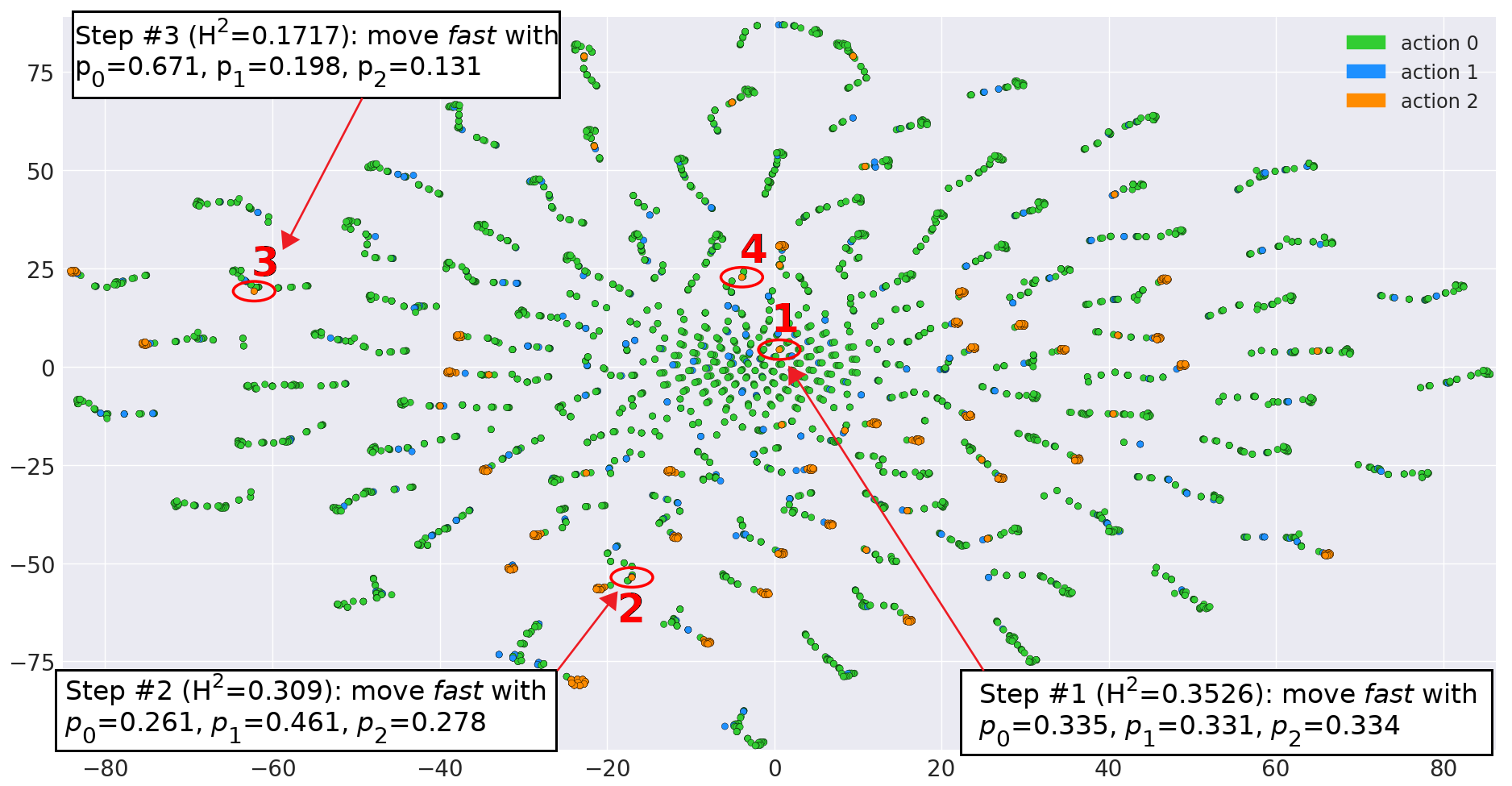} 
\caption{An higher resolution version of Figure~6 of the main paper} \label{fig:annotated_tsne}
\end{figure}

\begin{figure}[!ht]
\includegraphics[width=1.0\columnwidth]{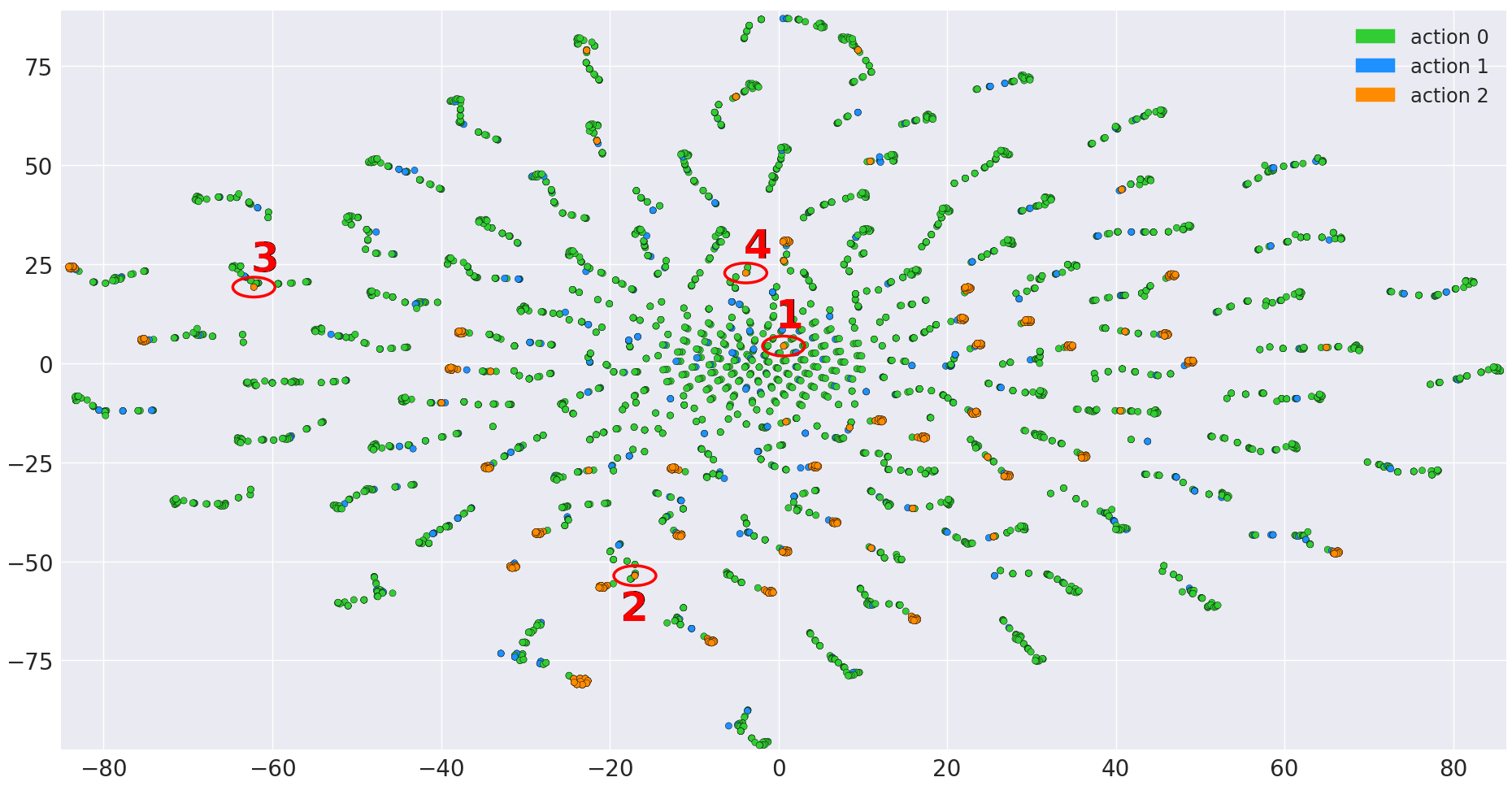} 
\caption{The t-SNE without any extra annotation} \label{fig:clean_tsne}
\end{figure}

In Section~4.3 of the main paper, we use \emph{T-distributed Stochastic Neighbor Embedding (t-SNE)} to present in a two dimensions image belief of the \emph{velocity regulation} problem (that is a probability distribution over $3^8$ states).
To improve the readability of this dense image, we provide a higher resolution version in Figure~\ref{fig:annotated_tsne}.
Figure~\ref{fig:clean_tsne} provides the same t-SNE but it does not contain any extra annotation.
We also include the two images as \emph{png} files in the supplementary material.

\section{XPOMCP implementation}
In the \texttt{code} folder of the supplementary material we provide an implementation of the XPOMCP methodology (in the \texttt{xpomcp} subforlder), a single trace that can be used to test XPOMCP (\texttt{example\_velreg\_trace}), and an implementation of pomcp that can be use to generate more traces (in subfolder \texttt{pomcp}).
The subfolders contain a \texttt{README.md} file that explains how to use them.

We presents the result of XPOMCP using the same rule presented in Section~4.3 of the main paper but a smaller trace. The ouput is:
\begin{lstlisting}[language=bash]
Import experiments
Find maximum number of satisfiable step in rule 0
fail to satisfy 3 steps out of 350
Check Formulas
rule: go at speed 2 if: P_0 >= 0.900 OR P_2 <= 0.017 OR
                       (P_0 >= 0.714 AND P_1 >= 0.113) 
Unsatisfiable steps:
ANOMALY: run example_velreg_trace/Run_5 step 0:
         action 2 with belief P_0 = 0.335 P_1 = 0.331 P_2 = 0.334
         --- Hellinger = 0.2735573417574736
         
run example_velreg_trace/Run_9 step 12:
action 2 with belief P_0 = 0.678 P_1 = 0.228 P_2 = 0.094
--- Hellinger = 0.029123804891139585

run example_velreg_trace/Run_1 step 26:
action 2 with belief P_0 = 0.880 P_1 = 0.099 P_2 = 0.022
--- Hellinger = 0.017102605324761978
\end{lstlisting}

The output report the number of steps that we fail to satisfy ($3$ out of $350$ in this case), the final \emph{rule} (i.e., an instantiation of the rule template provided in Section~4.3 of the main paper with the number of the trace), and a list of the unsatisfiable steps.
Only the first one is identified as an anomaly (using a threshold $\tau=0.1$ as in the main paper).